\documentclass[letterpaper, 10 pt, conference]{ieeeconf}  

\IEEEoverridecommandlockouts                              

\usepackage[pagebackref=true,breaklinks=true,letterpaper=true, citecolor=blue, colorlinks]{hyperref}
\usepackage{url}
\usepackage{cite}
\usepackage{graphicx}
\usepackage{graphics}
\usepackage{amsmath,amsfonts,amssymb}

\newcommand{\et}{{\em et al.\ }}

\newcommand{\RB}{\mbox{$\bf R$}}
\newcommand{\SB}{\mbox{$\bf S$}}

\newcommand{\XB}{\mbox{$\bf X$}}

\newcommand{\LC}{\mbox{$\mathcal L$}}

\newcommand{\NC}{\mbox{$\mathcal N$}}

\newcommand{\RC}{\mbox{$\mathcal R$}}

\newcommand{\beq}{\begin{equation}}
\newcommand{\eeq}{\end{equation}}
\newcommand{\bear}{\begin{eqnarray}}
\newcommand{\bears}{\begin{eqnarray*}}
\newcommand{\eear}{\end{eqnarray}}
\newcommand{\eears}{\end{eqnarray*}}
\newcommand{\bdm}{\begin{displaymath}}
\newcommand{\edm}{\end{displaymath}}
\newcommand{\lba}{\left[\begin{array}}
\newcommand{\ear}{\end{array}\right]}

\graphicspath{ {images/} }

\usepackage{tabularx}
\usepackage{multirow}
\usepackage[table]{xcolor}
\usepackage{longtable}
\usepackage{booktabs} 			
\usepackage{xcolor,colortbl}    

\title{Visual-Semantic Graph Attention Networks for \\ 
Human-Object Interaction Detection
}

\author{Zhijun Liang$^{1\dag}$, Juan Rojas$^{2\dag\ast}$, Junfa Liu$^{1}$, and Yisheng Guan$^{1\ast}$ 
 \thanks{$^{1}$The Biomimetic and Intelligent Robotics Lab (BIRL), School of Electromechanical Engineering, Guangdong University of Technology, 510006 Guangzhou, China. $^{2}$Dept. of Mechanical and Automation Engineering,  Chinese University of Hong Kong, Hong Kong, China.} 
 \thanks{$\dag$ Equal contribution. $\ast$ Corresponding authors (\url{ysguan@gdut.edu.cn} and \url{juan.rojas@cuhk.edu.cn}). The work in this paper is in part supported by the Frontier and Key Technology Innovation Special Funds of Guangdong Province (Grant No. 2017B050506008), the Key R\&D Program of Guangdong Province (Grant No. 2019B090915001).}
 }

\begin{document}

\maketitle
\thispagestyle{empty}
\pagestyle{empty}

\begin{abstract}

In scene understanding, robotics benefit from not only detecting individual scene instances but also from learning their possible interactions.
Human-Object Interaction (HOI) Detection infers the action predicate on a $<$human, predicate, object$>$ triplet. Contextual information has been found critical in inferring interactions. However, most works only use local features from single human-object pair for inference. Few works have studied the disambiguating contribution of subsidiary relations made available via graph networks. Similarly, few have learned to effectively leverage visual cues along with the intrinsic semantic regularities contained in HOIs.   
We contribute a dual-graph attention network that effectively aggregates contextual visual, spatial, and semantic information dynamically from primary human-object relations as well as subsidiary relations through attention mechanisms for strong disambiguating power.
We achieve comparable results on two benchmarks: V-COCO and HICO-DET.  
Code is available at \url{https://github.com/birlrobotics/vs-gats}.

\end{abstract}

\section{INTRODUCTION}\label{sec:intro}
\begin{figure}[t]
  \begin{center}
      \includegraphics[width=1\linewidth]{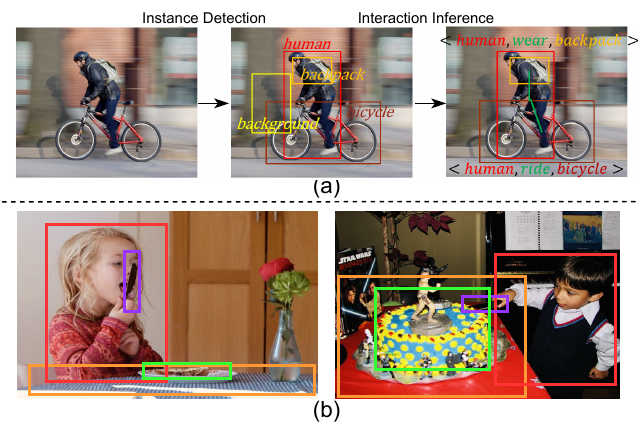}
  \end{center}
      \caption{
      (a) A general framework of HOI detection.
      (b) How subsidiary relations facilitate HOI detection: On the left, with the features from \textit{[human-knife]}, the model can easily infer \textit{``hold"} and \textit{``lick"} predicates for this tuple, while the spatial message from subsidiary relations \textit{[knife-table]} inhibits the model from choosing \textit{``cut\_with"}. On the right, if we just focus on the features from \textit{[human-cake]}, the model may output similar scores for the \textit{``cut''} and \textit{``light''} predicates since they share similar visual embedding features. However, messages from subsidiary relations \textit{[human-knife]} and \textit{[knife-cake]} promote $<$\textit{human,cut,cake}$>$.
      }
  \label{fig:introduce_picture}
\end{figure}

Human-Object Interaction (HOI) detection has recently gained important traction and has pushed forward robot's abilities to understand the visual world. Whilst computer vision has experienced extraordinary advances in object detection \cite{renNIPS15fasterrcnn,liu2016ssd,dai2016r}, human pose estimation \cite{Dabral2018LearningMotion,pavllo20193d} and scene segmentation \cite{he2017mask}; the harder problem of HOI detection has made less progress.
Generally, HOI detection starts with instance detection and continue with interaction inference as illustrated in Fig. \ref{fig:introduce_picture}(a). The goal is to infer an interaction predicate for the $<$human, predicate, object$>$ triplet. Note that humans can simultaneously interact with different objects and have different interactions with the same object. \textit{I.e.} for Fig. 1(b) on the right, the HOIs could be $<$\textit{human,cut,cake}$>$, $<$\textit{human,hold,knife}$>$ and $<$\textit{human,cut\_with,knife}$>$. Therefore, HOI detection is a multi-label problem, which requires better understanding of contextual information for better inference. 

Over time, researchers have exploited a variety of contextual cues including visual, spatial, semantic, human pose to better understand a scene \cite{chao2018learning, gkioxari2018detecting, gao2018ican, Li_2019_CVPR, xu2019learning, bansal2019detecting, wan2019pose, gupta2018nofrills, 2019PeyreDetecting}. Researchers have also used a variety of architectures (Sec. \ref{sec:relatedWorks}). But most works have only leveraged local-primary relations in the scene to infer interactions. Very recently graph attention nets \cite{qi2018learning} have been considered. However, they just use limited contextual cues in a signal graph with complicated features updating mechanism. 

Under a graph-based structure, image instance proposals yield graph nodes connected by edges. A primary relation is defined as the immediate human-object relation under consideration; whilst subsidiary relations are all other connections in the graph. In this manner, primary and subsidiary relations are relative. One key insight of this work is that leveraging various contextual cues from subsidiary relations aid to disambiguate in HOI detection. 
For example, in Fig. \ref{fig:introduce_picture} consider the \textit{[human-knife]} the primary relation on the left and the \textit{[human-cake]} the primary relation on the right. On the left, this primary relation's visual and spatial cues might predict \textit{``hold''} and \textit{``cut\_with''}. But spatial cues from the subsidiary relations \textit{[knife-bread]} inhibit the system from choosing \textit{``cut\_with''}. On the right, the primary relation's cues might predict \textit{``cut''} or \textit{``light''} as these actions share similar visual embeddings. However, only when the system pays attention to the \textit{[knife-cake]} and \textit{[human-knife]} subsidiary contextual cues can it easily infer that \textit{``cut''} is the right interaction.
Another key insight of our work is that HOIs also posses intrinsic semantic regularities that aid detection despite diverse scenes. For instance, semantic cues from \textit{human} and \textit{knife}, may help the model focus on the actions related to the \textit{knife} instead of actions like \textit{``ride''}.

\begin{figure*}
    \centering
    \includegraphics[width=1\linewidth]{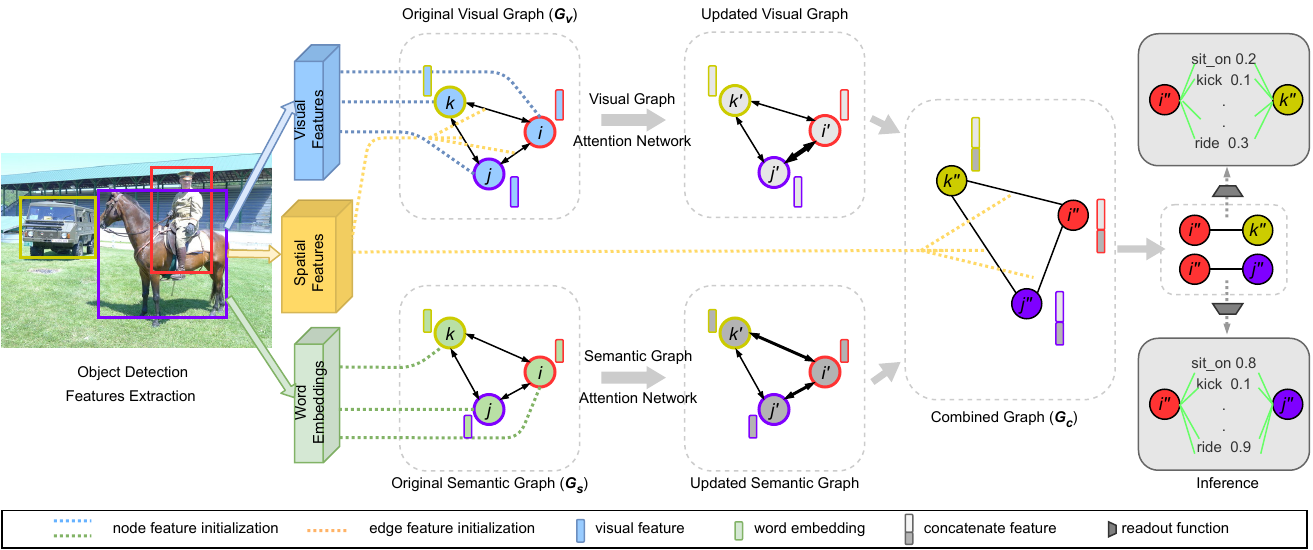}
    \caption{Visual-Semantic Graph Attention Network: After instance detection, a visual-spatial graph and a semantic graph are created. Node features are dynamically updated through graph attention networks (Sec. \ref{subsec:GATs}). We combine these updated graphs and then perform a readout step on box-pairs to infer all possible predicates between one human and one object.}
    \label{fig:graph_inference}
\end{figure*}

In this paper, we study the disambiguating power of subsidiary scene relations and intrinsic semantic regularities via a double graph attention network that aggregates visual-spatial and semantic information in parallel. This graph-based attention network structure explicitly enables the model to leverage rich information by integrating and broadcasting information through graph structure with attention mechanism. We call our system: Visual-Semantic Graph Attention Networks (VS-GATs).
As shown in Fig. \ref{fig:graph_inference}, our method begins by using instance detection to yield bounding-boxes with visual features and semantic categories. From this, a pair of scene graph are created. 
The first graph's nodes are instantiated from the bounding-box visual features; while the edges are instantiated from corresponding spatial features.
The second graph's nodes are instantiated from corresponding word embedding features. 
Two graph attention networks then update the node features of each graph via message passing.
A combined graph is created by concatenating both graph's node updated features. Then inference is done through a readout step on paired human-object nodes. Please see Sec. \ref{sec:vsgat} for more details.

On HICO-DET dataset, our method achieves comparable results for the Full, Rare and Non-Rare categories with mAP of \textbf{20.27}, \textbf{16.03} and \textbf{21.54} respectively. On  V-COCO dataset, our model also obtains promising results with mAP of \textbf{50.6}.

\section{RELATED WORK}\label{sec:relatedWorks}
In this section, we present the related works by keying in on the architecture type: multi-streams DNN and GNN.
\paragraph{\textbf{Multi DNN Streams with Various Contextual Cues}}
A primary way to do HOI detection has been to extract visual features from instance detectors along with spatial information to instantiate multi-streams of DNNs. Each stream contains detected human, object, and other contextual features. A final fusion step is designed for inference. \cite{chao2018learning,gkioxari2018detecting,gao2018ican}.
Lu \et \cite{lu2016visual} considered semantic information under the multi-stream DNN setting stating that interaction relationships are also semantically related to each other.
Peyre \et \cite{2019PeyreDetecting} used a concept of visual analogies. They instantiated a stream using a visual-semantic embedding of the triplet resulting in a trigram.
Gupta \et \cite{gupta2018nofrills} and Li \et \cite{Li_2019_CVPR} used fine-grained layouts of the human pose and leverage relation elimination or interactiveness modules to improve inference. Wan \et \cite{wan2019pose} further considered not only human pose but also human body part features to enhance inference. 
Recently, some works have been parallelly developed with ours. Hou \et \cite{houECCV2020visualCompositional} proposed a framework to perform compositional learning by sharing visual components across images for new interaction samples generation. Kim \et \cite{kimECCV2020Co-occurrence} extended \cite{gupta2018nofrills} by explicitly leveraging the action co-occurrence priors for HOI detection. Li \et further explored detailed 2D-3D joint representation \cite{LiCVPR2020detailed3d} and detailed human body part states \cite{LiCVPR2020PaStaNet} for better HOI detection.
\textbf{However}, these works are limited to local features for inference, which do not consider subsidiary relations. In this work, we explore using graph structure network to take the subsidiary relations into account for learning rich contextual information to facilitate HOI detection. 

\paragraph{\textbf{Graph Neural Networks}}
GNNs have been used to model scene relations and knowledge structures.
Yang \et \cite{yang2018graph} proposed an attentional graph convolution network to aggregate global context for scene graph generation. 
Sun \et \cite{sun2019relational}, do multi-person action forecasting in video. They use a recursive GNN on visual and spatio-temporal features to update the graph. 
Kato \et \cite{kato2018compositional} use an architecture that consists of one stream of convolutional features and another stream composed of a semantic graph for HOI classification. 
Leaning on the concept of semantic regularities, Xu \et \cite{xu2019learning} similarly use a visual stream with convolutional features for human and object instances and a parallel knowledge graph for HOI detection. 

 To data, only Qi \et \cite{qi2018learning} have used GAT architecture for HOI detection. Their method (GPNN) creates nodes and edges from visual features. The graph structure is set by an adjacency matrix and features is updated by a weighted sum of the messages of the other nodes. Finally, a node readout function is used for interaction inference. 
Our method is similar, but different. \textbf{First}, as illustrated in Fig. \ref{fig:graph_inference}, instead of using single graph our model uses a novel parallel dual-attention graph architecture which also takes semantic cues into account. Furthermore, we identify spatial features as critical in the final inference step. \textbf{Second}, we leverage a simpler but more effective node features updating mechanism without multiple iterations using in \cite{qi2018learning}. A \textbf{final} difference is that in \cite{qi2018learning}, Qi \et use a node readout function to \textit{separately} infer actions for each node. We find it more reasonable to jointly infer actions with the combined features of the human and object; as such, we use an edge readout function (Eqtn. \ref{eqtn:actionScore}) to infer the interaction from the edges connected to the human. Overall, our model outperforms GPNN by a great margin on both datasets.

\section{VISUAL-SEMANTIC GRAPH ATTENTION NETWORK}\label{sec:vsgat}  
In this section,  we first define graphs and then describe the contextual features and our VS-GATs. 
\subsection{Graphs}\label{subsec:graphs} 
A graph $G$ is defined as $G=(V, E)$, where $V$ is a set of $n$ nodes and $E$ is a set of $m$ edges. Node features and edge features are denoted by $\mathbf{h}_v$ and $\mathbf{h}_e$ respectively. Let $v_i \in V$  be the $ith$ node and $e_{i,j}=(v_i,v_j) \in E$ be the directed edge from $v_i$ to $v_j$. 
A graph with $n$ nodes and $m$ edges has a node features matrix $\XB_v \in \RC^{n \times d}$ and an edge feature matrix $ \XB_e \in \RC^{m \times c} $ where $\mathbf{h}_{v_i} \in \RC^d$ is the feature vector of node $i$ and $\mathbf{h}_{e_{i,j}} \in \RC^c $ is the feature vector of edge $(i,j)$. Fully connected edges imply $e_{i,j} \neq e_{j,i}$. 
\subsection{Contextual Features} \label{subsec:relatedFeatures}
\paragraph{\textbf{Visual Features}} 
Visual features are extracted from human and object proposals generated from Faster R-CNN \cite{renNIPS15fasterrcnn}. As with \cite{gupta2018nofrills}, first, the RPN generates human and object proposals. Thus, for an image $I$, the $i$th human bounding-box $b_h^i$ and the $j$th object bounding-box $b_o^j$ are used to extract latent features from Faster R-CNNs last fully-connected layer ($FC7$ after the ROI pooling layer) to instantiate the visual graph ($G_v$) nodes as illustrated in Fig. \ref{fig:graph_inference}.

\paragraph{\textbf{Spatial Features}} 
Instance spatial features such as bounding box locations and relative locations are informative about the relationship that proposals have with each other \cite{Zhuang_2017_ICCV,hu2017modeling,plummer2017phrase,zhang2017visual}. Consider the ``ride'' predicate, we can deduce that human is above the object.

Given a pair of bounding boxes, their paired-coordinates are given by  $(x_i,y_i,x_j,y_j)$ and $(x_i^{'},y_i^{'},x_j^{'},y_j^{'})$ and centres are denoted as $(x_c,y_c)$ and $(x_c^{'},y_c^{'})$. Along with respective areas $A$ and $A^{'}$ and an image area $A^{I}$ of size $(W,H)$. 

Instance spatial features $\mathbf{h}_{f_{ij}}=\mathbf{h}_{f_{ij}^{rs}} \bigcup \mathbf{h}_{f_{ij}^{rp}} $ can be grouped into relative scale features 
$\mathbf{h}_{f_{ij}^{rs}}= \left[ \frac{x_i}{W}, \frac{y_i}{H}, \frac{x_j}{W}, \frac{y_j}{H}, \frac{A}{A^I} \right]$, and relative position features $\mathbf{h}_{f_{ij}^{rp}}= 
        \left[ (\frac{x_i-x_i^{'}}{x_j^{'}-x_i^{'}}), (\frac{y_i-y_i^{'}}{y_j^{'}-y_i^{'}}), \log(\frac{x_j-x_i}{x_j^{'}-x_i^{'}}), \log(\frac{y_j-y_i}{y_j^{'}-y_i^{'}}), 
        \frac{x_c-x_c^{'}}{W}, \frac{y_c-y_c^{'}}{H} \right].$
Spatial features are used to: (i) instantiate the edges in the Visual graph ($G_v$) (ii) and in the Combined Graph ($G_c$) as illustrated in Fig. \ref{fig:graph_inference}.

\paragraph{\textbf{Semantic Features}}
In this work, we use Word2vec embeddings \cite{mikolov2013distributed} as semantic features.  
We use the publicly available Word2vec vectors pre-trained on the Google News dataset (about 100 billion words) \cite{Google2013GoogleHosting.}. 
All existing object classes in the dataset are used to obtain the 300-dimensional Word2vec latent vector representations offline.
These semantic features are used to instantiate the nodes in the semantic graph ($G_s$) as illustrated in Fig. \ref{fig:graph_inference}.
\subsection{Graph Attention Networks} \label{subsec:GATs}
In graph neural networks, a node's features are updated by aggregating its neighboring nodes' features. 
The node updated features $\tilde{\mathbf{h}}_{v_i}$ for node $v_i$ are generically defined as:
\begin{eqnarray}
   \mathbf{a}_{v_i}=f_{aggregate}  
   \left(  
      \left\{ 
         \mathbf{h}_{v_j} : v_j \in \NC_i  
      \right\}
   \right) \\
    \tilde{\mathbf{h}}_{v_i} = f_{update}
    \left( 
      \mathbf{h}_{v_i},\mathbf{a}_{v_i}
    \right).
\end{eqnarray}
Where $\NC_i$ is the set of nodes adjacent to $v_i$. Also, the common $f_{aggregate}(\cdot)$ is averaging: 
\begin{eqnarray}
   \label{eqtn:aver}
   \mathbf{a}_{v_i}= 
         \frac{1}{|\NC_i|} \sum_{v_j \in \NC_i} \mathbf{h}_{v_j}. 
\end{eqnarray}
\subsubsection{Visual Graph Attention Network}\label{subsubsec:VGAT}
The visual graph is constructed with visual features and instance spatial features illustrated in Sec. \ref{subsec:relatedFeatures}. We first use an edge function $f_{edge}(\cdot)$ to \textit{encode the relation features between two connected nodes}:
\begin{eqnarray}
    \mathbf{h}_{e_{ij}}=f_{edge}([\mathbf{h}_{v_i},\mathbf{h}_{f_{ij}},\mathbf{h}_{v_j}]).
\end{eqnarray}

Note that in the first step, the object detector may yield irrelevant instances, then using Eqtn. \ref{eqtn:aver} for node features aggregation might introduces significant noise. Instead, we leverage an attention mechanism to mitigate this problem:
\begin{eqnarray}
    \alpha_{ij} = \frac{exp(f_{attn}( \mathbf{h}_{e_{ij}} ))}{ \sum_{v_o \in \NC_i}{exp(f_{attn}( \mathbf{h}_{e_{io}}))}}.
\end{eqnarray}
where $\alpha_{ij}$ is the soft weight indicated the importance of node $v_j$ to node $v_i$ via this softmax operation.
Then we apply a weighted sum and use the updated function $f_{update}(\cdot)$ to update each node's features:
\begin{eqnarray}
    \mathbf{z}_{v_i} = \sum_{v_j \in \NC_i} \alpha_{ij} ( \mathbf{h}_{v_j} + \mathbf{h}_{e_{ij}}) \\
   \tilde{ \mathbf{h} }_{v_i} = f_{update}([ \mathbf{h}_{v_i}, \mathbf{z}_{v_i}]).
\end{eqnarray}
\textit{Note that $\mathbf{z}_{v_i}$ consists of the accumulated latent relation features of all the neighboring node connected to $v_i$}. 

At this point, we can get an ``updated visual graph'' with new features as illustrated  in Fig. \ref{fig:graph_inference}. The different edge thickness' represent the soft weight distributions. 
Note that in our method, we implement $f_{attn}(\cdot)$, $f_{update}(\cdot)$, and $f_{edge}(\cdot)$ as a single fully-connected layer network with node dimensions of 1, 1024, and 1024 respectively. 
\subsubsection{Semantic Graph Attention Network}\label{subsubsec:SGAT}
In the semantic graph, Word2vec latent representations of the class labels of detected objects are used to instantiate the graph's nodes. We denote $\mathbf{w}_{v_i}$ as the word embedding for node $i$.
As with the visual graph, we use an $f_{edge}^\prime(\cdot)$ function and an $f_{attn}^\prime(\cdot)$ function to compute the 
distributions of soft weights $\alpha_{ij}^\prime$ on each edge 
$
   \alpha_{ij}^\prime = softmax(f_{attn}^\prime(f_{edge}^\prime([ \mathbf{w}_{v_i}, \mathbf{w}_{v_j}]))).
$
Then, the global semantic features for each node are computed through the linear weighted sum:
\begin{equation}
   \mathbf{z}_{v_i}^{'} = \sum_{v_j \in N_i} \alpha_{ij}^\prime \mathbf{w}_{v_j}. 
\end{equation}
After that, we update the node's features as
$
   \tilde{ \mathbf{w} }_{v_i} = f_{update}^\prime([ \mathbf{w}_{v_i}, \mathbf{z}_{v_i}^{'} ]).
$
As with the visual graph, we output an ``updated visual graph'' with new features as shown in Fig. \ref{fig:graph_inference}. Similarly, $f_{edge}^\prime(\cdot)$, $f_{attn}^\prime(\cdot)$, 
and $f_{update}^\prime(\cdot)$ are designed in the same way as with the visual graph.
\subsection{Combined Graph.} \label{subsec:fusion_module}
To jointly leverage the dynamic information of both the visual ($G_v$) and the semantic ($G_s$) GATs, it is necessary to fuse them as illustrated in the ``Combined Graph'' ($G_c$) of Fig. \ref{fig:graph_inference}. We concatenate the features of each of the updated nodes to produce new nodes and initialize the edges with the original $\mathbf{h}_f$ described in Sec. \ref{subsec:relatedFeatures}. We denote the combined node features as $\mathbf{h}_{v_i}^c$ for node $i$, where ${\mathbf{h}}_{v_i}^c=[\tilde{ \mathbf{h}}_{v_i}, \tilde{ \mathbf{w}}_{v_i}]$.
\subsection{Readout and Inference.}\label{subsec:readOut}
Through above graph attention networks, \textit{I.e.} for Fig. \ref{fig:introduce_picture} (b) on the right, human combined node features have encoded relation cues of [human, cake], [human, knife] and [human, table], while knife's have encoded relation cues of [knife, cake], [knife, table] and [knife, human]. 
Then we box-pair all specific human-object as illustrated in Fig. \ref{fig:graph_inference}. \textit{In doing so, when predicting the relation between human and cake, the model can not only leverage the cues of [human, cake] but other subsidiary cues from [human, cake], [cake, knife]. }

With box-pairing, we finally construct the concatenated representation $\zeta_{ij}=[ \mathbf{h}_{v_i}^c, \mathbf{h}_{f_{ij}}, \mathbf{h}_{v_j}^c ]$ for prediction. To compute the action category score $\mathbf{s}^a \in \RC^k $, where \textit{k} denotes the total number of possible actions, we apply an edge readout function $f_{readout}(\cdot)$ \footnote{A multi-layer perceptron with 2 layers of dimensions 1024 and 117 for HICO-DET, and 1024 and 24 for V-COCO.},
and then apply a binary sigmoid classifier for each action category:
\begin{equation}
    \label{eqtn:actionScore}
    \mathbf{s}^a = sigmoid( f_{readout}(\zeta) ).
\end{equation}
The final score of a triplet's predicate  $\SB_{\RB}$ can be computed through the chain multiplication of the action score $\mathbf{s}^a$, the detected human score $s_h$ and the detected object score $s_o$ from object detection as:
$
    \SB_{\RB} = s_h * s_o * \mathbf{s}^a.
$
\hfill\newline\newline 
\noindent \textbf{Training.}\label{subsec:training}
For training, we use a multi-class cross-entropy loss that is minimized between action scores and the ground truth action label:
\begin{equation}
    \LC = \frac{1}{N \times k} \sum_{i=1}^N \sum_{j=1}^k  BCE(s_{ij}, y_{ij}^{label})
\end{equation}
where $N$ is the number of all box-pairs in each mini-batch and $s_{ij} \in \mathbf{s}^a_i$. See Sec. \ref{subsec:experimental_setup} for more training details.

\section{EXPERIMENTS AND RESULTS}\label{sec:exp_results}
%
\subsection{Experimental Setup}\label{subsec:experimental_setup}
\noindent\textbf{Datasets.} In this work, we use two common benchmarks: V-COCO \cite{DBLP:GuptaM15} and HICO-DET \cite{chao2018learning}. 
V-COCO has 2,533, 2,867, 4,946 training, validating, and testing images respectively and 16,199 human instances annotated with 26 action categories.
Compared to V-COCO, HICO-DET is much larger and diverse. HICO-DET has 38,118 and 9,658 training and testing images. The 117 interaction classes and 80 objects yield 600 HOI total categories. There are 150K annotated human-object pair instances. HICO-DET is divided in three classes: Full: all 600 categories; Rare: 138 categories with less than 10 training instances, and Non-Rare: 462 categories.

\noindent \textbf{Evaluation Metrics.} As with prior works, we use the standard mean average precision (mAP) metric to evaluate the model's detection performance. 
In this case, we consider a detected result with the form $<$human, predicate, object$>$ is positive when the predicted verb is true and both the detected human and object bounding boxes have the intersection-of-union (IoU) exceed 0.5 with respect to the corresponding ground truth.

\noindent \textbf{Implementation Details.}\label{subsec:impldetails} Our architecture is built on Pytorch and the DGL library \cite{DGL2019}. For object detection we use Pytorch's re-implemented Faster R-CNN API \cite{renNIPS15fasterrcnn}. Faster R-CNN use a ResNet-50-FPN backbone \cite{he2016deep,lin2017feature} trained on the COCO dataset\cite{Lin2014MicrosoftContext}. The object detector and Word2vec vectors are frozen during training. We keep the human bounding-boxes whose detection score exceeds 0.8, while for objects we use a 0.3 score threshold. 


All neural network layers in VS-GAT are constructed as MLPs as mentioned in previons. Training on HICO-DET, we use batch size of 32 and a dropout rate of 0.3. 
We use an Adam optimizer with a initial learning rate of 1e-5. We reduce the learning rate to 3e-6 at 200 epochs and stop training at 250 epochs.
As for the activation function, we use a LeakyReLU in all attention network layers and a ReLU elsewhere.
For V-COCO dataset, we train the model with the same hyperparameters except for the dropout rate (from 0.3 to 0.5) and the training epoch (from 250 to 600).

\begin{table}[tb]
\centering
    \caption{mAP performance comparison on HICO-DET test set.}
    \label{tbl:hico_mAP}
    \begin{tabular}{lccc}
    \toprule
    Method                   & Full(600)$\uparrow$   & Rare(138)$\uparrow$   & Non-Rare(462)$\uparrow$  \\
    \bottomrule
    InteractNet \cite{gkioxari2018detecting}              & 9.94  & 7.16  & 10.77    \\
    GPNN \cite{qi2018learning}                            & 13.11 & 9.34  & 14.23    \\
    iCAN \cite{gao2018ican}                               & 14.84 & 10.45 & 16.15    \\
    Xu \textit{et al.} \cite{xu2019learning}              & 14.70  & 13.26 & 15.13    \\
    Bansal \textit{et al.} \cite{bansal2019detecting}       & 16.96  & 11.73  & 18.52     \\
    Gupta \textit{et al.} \cite{gupta2018nofrills}        & 17.18 & 12.17 & 18.68    \\
    Li \textit{et al.} \cite{Li_2019_CVPR}                      & 17.22 & 13.51 & 18.32    \\
    PMFNet \cite{wan2019pose}                             & 17.46 & 15.65 & 18.00       \\
    Peyre \textit{et al.} \cite{2019PeyreDetecting}       & 19.40  & 14.60  & 20.90     \\
    VCL \cite{houECCV2020visualCompositional}       & 19.43  & 16.55  & 20.29     \\
    PaStaNet*-Linear \cite{LiCVPR2020PaStaNet}       & 19.52  & 17.29  & 20.19     \\
    ACP \cite{kimECCV2020Co-occurrence}       & 20.59  & 15.92  & 21.98     \\
    DJ-RN \cite{LiCVPR2020detailed3d}       & 21.34  & 18.53  & 22.18     \\
    \toprule
    \textbf{Ours(VS-GATs)}       & 20.27$\pm0.10$ & 16.03$\pm0.42$ & 21.54$\pm0.02$     \\
    \bottomrule
    \end{tabular}
\end{table}
\begin{figure*}[t]
    \centering
    \includegraphics[width=1\linewidth]{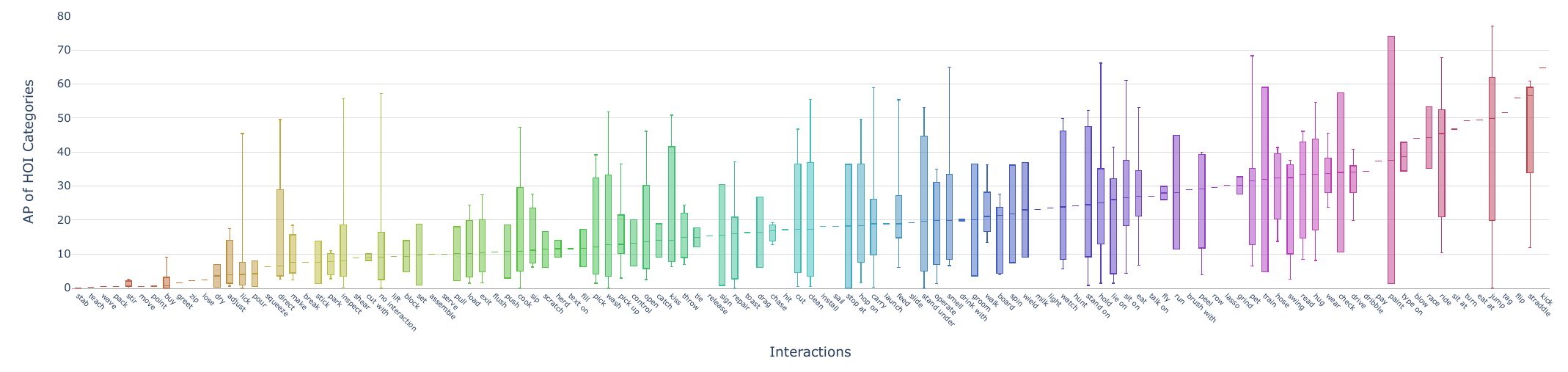}
    \caption{Spread of performance across objects for a given interaction on HICO-DET. The horizontal axis is sorted by median AP.}
    \label{fig:perfDist}
\end{figure*}
\subsection{Results}\label{subsec:result}
\subsubsection{Quantitative Results and Comparisons}
Our experiments show our model achieves the promising results on both HICO-DET and V-COCO. 
For fair comparison, results of all compared baselines are chosen with their object detectors trained on COCO dataset. 

On HICO-DET, we train our model three times to get the more convincing results. Our model obtains comparable results in all three categories with map of 20.27, 16.03, 21.54 respectively. We surpass GPNN that also uses GNN by 7.16 mAP gain. We also outperform most of multi-streams works \cite{gkioxari2018detecting, gao2018ican, xu2019learning, bansal2019detecting, gupta2018nofrills, Li_2019_CVPR, wan2019pose, 2019PeyreDetecting, houECCV2020visualCompositional, LiCVPR2020PaStaNet} including works that leveraged human pose \cite{gupta2018nofrills, Li_2019_CVPR, wan2019pose, LiCVPR2020PaStaNet} except \cite{kimECCV2020Co-occurrence} and \cite{LiCVPR2020detailed3d}. However, DJ-RN \cite{LiCVPR2020detailed3d} is resource-consuming for further leveraging the expensive 2D even 3D pose estimator for detailed joint representation. And ACP \cite{kimECCV2020Co-occurrence} is labor-intensive for explicitly constructing the action co-occurrence prior which is various in different datasets. Meanwhile, it also inserts the pose detector. However, our method tries to learn co-occurrence relationship implicitly by data-driven, and also demonstrates better disambiguating power with 20.27 mAP favorable result even without extra pose cues.

\begin{table}[h]
\centering
    \caption{mAP performance comparison on V-COCO test set.}
    \label{tbl:vcoco_map}
    \begin{tabular}{lc}
    \toprule
    Method              & $AP_{role}$ (Scenario 1)  \\
    \midrule
    Gupta \textit{et al.} \cite{DBLP:GuptaM15}      & 31.8 \\
    InteractNet \cite{gkioxari2018detecting}        & 40.0  \\
    GPNN \cite{qi2018learning}                      & 44.0     \\
    iCAN \cite{gao2018ican}                         & 45.3     \\
    Xu \textit{et al.} \cite{xu2019learning}        & 45.9    \\
    Li \textit{et al.} \cite{Li_2019_CVPR}          & 47.8     \\
    VCL \cite{houECCV2020visualCompositional}       & 48.3    \\
    PMFNet \cite{wan2019pose}                   & 52.0      \\
    PMFNet (w/o pose) \cite{wan2019pose}          & 48.6      \\
    ACP \cite{kimECCV2020Co-occurrence}       & 53.0    \\
    \midrule
    \textbf{Ours (VS-GATs)}                     & 50.6  \\
    \bottomrule
    \end{tabular}
\end{table}


On V-COCO, VS-GATs achieves 50.6 mAP which exceeds GPNN by 6.6 mAP and also outperforms most of other STOAs using multi-streams network except \cite{wan2019pose} and \cite{LiCVPR2020detailed3d}. \cite{wan2019pose} develops a well-defined \textit{Zoom-in Module} which utilizes \textit{fine-grained human pose} as well as \textit{body part features} to extract detailed local appearance cues, which makes their model surpass \cite{Li_2019_CVPR} by a great margin on the small-scale dataset. However, \cite{wan2019pose} and \cite{Li_2019_CVPR} have a similar performance on the more diverse dataset HICO-DET. Without the \textit{Zoom-in Module}, \cite{wan2019pose} obtains the result of 48.6, which worse than our model.  ACP \cite{kimECCV2020Co-occurrence} which also use pose cues set up a new STOA result 53.0 mAP, which indicates explicitly leveraging the co-occurrence prior is more effective on the small-scale dataset than all other methods that explore this cues implicitly.

In Fig. \ref{fig:perfDist}, we also visualize the performance distribution of our model across objects for a given interaction. As mentioned in \cite{gupta2018nofrills}, it still holds that interactions that occur with just a single object (e.g. 'kick ball' or 'flip skateboard') are easier to detect than those predicates that interact with various objects. Compared to \cite{gupta2018nofrills}, the median AP of those interactions like 'cut' and 'clean' shown in Fig. \ref{fig:perfDist} outperform those in \cite{gupta2018nofrills} by a considerable margin. We hold that gains from our works are due to learning the rich contextual relationship by the dual attention graphs which enable each node to leverage contextual cues from a wide-spread set of (primary and subsidiary) other nodes as illustrated in \ref{subsec:readOut}.  

By comparation, we note that human pose cues are informative for HOI inferring \cite{wan2019pose, LiCVPR2020detailed3d, kimECCV2020Co-occurrence} and the object detector also play a important role in HOI detection \cite{bansal2019detecting, houECCV2020visualCompositional}. We will try to export those in our future works.  

\subsubsection{Qualitative Results}
\begin{figure*}[t]
    \centering
    \includegraphics[width=1\linewidth]{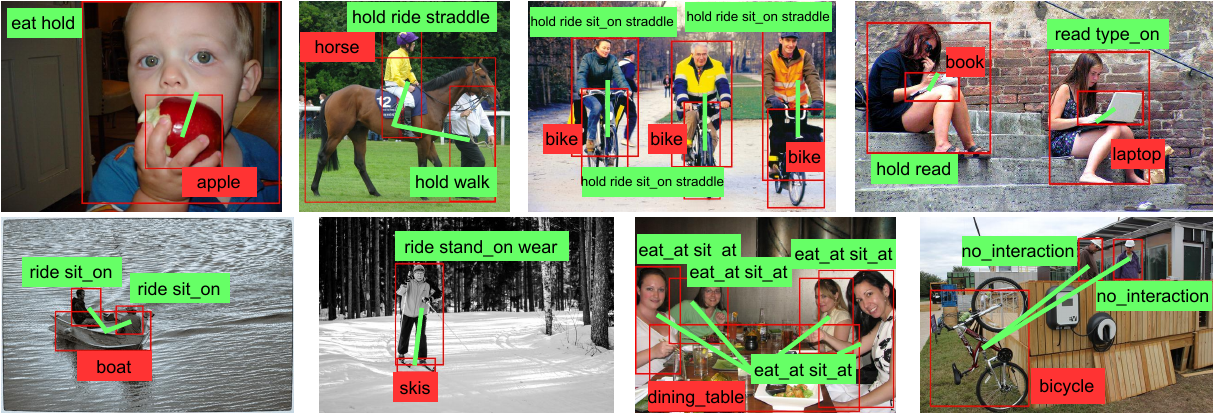}
    \caption{HOI detection examples on the HICO-DET testing dataset. Humans and objects are shown in red bounding boxes. Interaction classes are shown on the human bounding box. Interactive objects are linked with the green line. We show all triplets whose inferred \textit{action score} (\ref{eqtn:actionScore}) is greater than 0.3.}
    \label{fig:qualitative}
\end{figure*}

Fig. \ref{fig:qualitative} shows some $<$human, predicate, object$>$ triplets' detection results on HICO-DET test set. From the results, our proposed model is able to detect various kinds of HOIs such as: single person-single object, multi person-same object, and multi person-multi objects. 

\begin{table}[h]
\centering
    \caption{mAP performance for various ablation studies.}
    \label{tbl:ablationTbl}
    \begin{tabular}{lccc}
    Method                                      & Full$\uparrow$  & Rare$\uparrow$  & Non-Rare$\uparrow$ \\
    \midrule
    \textbf{Ours(VS-GATs)}                      & \textbf{20.27}$\pm0.10$ & \textbf{16.03}$\pm0.42$ & \textbf{21.54}$\pm0.02$  \\
    \midrule
    01 $G_V$ only                                  & 19.68$\pm0.05$ & 15.42$\pm0.25$ & 20.96$\pm0.03$     \\
    02 $G_S$ only                                 & 14.34$\pm0.02$ & 10.91$\pm0.16$ & 15.36$\pm0.03$     \\
    \midrule
    03 w/o attention                              & 19.75$\pm0.05$ & 14.84$\pm0.31$ & 21.21$\pm0.05$     \\
    04 w/o $\mathbf{h}_f$ in $G_C$              & 19.32$\pm0.06$ & 15.54$\pm0.15$ & 20.45$\pm0.06$     \\
    \midrule
    05 Message passing in $G_C$                   & 20.04$\pm0.14$ & 15.24$\pm0.22$ & 21.47$\pm0.18$     \\
    06 Unified V-S graph                          & 19.92$\pm0.07$ & 15.46$\pm0.66$ & 21.25$\pm0.12$     \\
    \bottomrule
    \end{tabular}
\end{table}


\subsection{Ablation Studies} \label{subsec:ablationStudies}
In this section, we choose HICO-DET as training dataset to study the impact of each component in our model. We conduct the following six tests (we run each test three times to get the results): 
\\ \noindent \textbf{01 Visual Graph Only: $G_V$ only.} 
In this test, we remove the Semantic-GAT and keep the Visual-GAT, attention, and inference the same. This study will show the importance of leveraging the semantic cues.
\\ \noindent \textbf{02 Semantic Graph Only: $G_S$ only.} 
In this test, we remove the Visual-GAT and keep the Semantic-GAT, attention, and inference the same. This study will show the importance of aggregating visual and spatial cues. 
\\ \noindent \textbf{03 Without Attention.} 
In this test, we use the averaging mechanism (Eqtn. \ref{eqtn:aver}) to aggregate features instead of the weighted sum attention mechanism.
\\ \noindent \textbf{04 Without Spatial Features ($\mathbf{h}_f$) in $G_C$.} 
In this test, we remove spatial features from the edges of the combined graph $G_C$ to study the role that spatial features can play after the aggregation of features across nodes.
\\ \noindent \textbf{05 Message Passing in $G_C$. } 
In this test, we leverage an additional graph attention network to process the combined graph which is similar to what we do to the original visual-spatial graph. We examine if there would be a gain from an additional message passing on $G_C$ with combined feature from $G_V$ and $G_S$.
\\ \noindent \textbf{06 Unified V-S Graph.} 
In this test, we choose to start with a single graph in which visual and semantic features are concatenated in the nodes from the start. Spatial features are still used to instantiate edges. This test examines if there would be a gain from using combined visual-semantic features from the start instead of through separate streams.

We now report on the ablation test results. For the Full category, study 01 yields an mAP of 19.68 which is a large portion of our full model mAP result suggesting that semantic cues can promote HOI detection but less effective than visual cues. When only using the Semantic graph in test 02, the effect is less marked, which also indicates the visual and spatial features play a primary role in inferring HOI. When combining these 3 contextual cues in a graph but not using the attention mechanism in test 03, the map results drop to 19.75. This suggests that attention mechanism assists the model for better HOI detection. Afterwards, inserting attention but removing spatial features at the end in test 04 hurts. This indicates that spatial features, even after the aggregation stage, are helpful. By inserting spatial features in the combined graph we are basically using a skip connection step in neural networks which has also shown to help classification. In test 05, we learn that additional message passing in the combined graph does not confer additional benefits. Similarly with test 06, a combined V-S graph is still not as effective as separating cues early on. This suggests that visual cues and semantic cues may have some degree of orthogonality to them even though they are related to each other. 

\section{CONCLUSION}\label{sec:conclusion}
In this paper we present a novel HOI detection architecture that studies and leverages the role of not only primary human-object contextual cues in interaction, but also the role of subsidiary relations. We show that multi-modal contextual cues from visual, semantics, and spatial data can be graphically represented through graph attention networks to leverage primary and subsidiary contextual relations. Our work have a promising results on both HICO-DET and V-COCO dataset.





\bibliographystyle{IEEEtran}
\bibliography{references}

\end{document}